\documentclass[runningheads]{llncs}
\usepackage{acronym}
\usepackage{xspace}
\usepackage{xcolor}
\usepackage{amssymb}
\usepackage{tcolorbox}
\usepackage{url}
\newtcolorbox{andreabox}{
  colback=gray!15,    
  colframe=blue!60,   
  boxrule=0.5pt,      
  arc=2mm,            
  left=1mm,           
  right=1mm,
  top=1mm,
  bottom=1mm
}

\newcommand{\dirty}{{Dirtify \xspace}}
\newcommand{\ml}{{Machine Learning \xspace}}

\usepackage[T1]{fontenc}
\usepackage{graphicx}
\usepackage{subcaption}
\begin{document}
\title{Measuring the Sensitivity of Classification Models with the Error Sensitivity Profile}
\titlerunning{Measuring the Sensitivity of Classification...}
%
\author{Andrea Maurino\inst{1}\orcidID{0000−0001−9803−3668]} }
\authorrunning{A. Maurino}
%
\institute{Università degli studi di Milano Bicocca \\
Dipartimento di Informatica, Sistemistica e Comunicazione\\
Viale Sarca 336,20126 Milano, Italy\\
\email{andrea.maurino@unimib.it}}
\maketitle              
\begin{abstract}

The quality of training data is critical to the performance of machine learning models. In this paper, the Error Sensitivity Profile (ESP) is proposed. It quantifies the sensitivity of model performance to errors in a single feature or in multiple features. By leveraging ESP, data-cleaning efforts can be prioritized based on error types and features most likely to affect model performance. To support the computation of this metric, an integrated suite of tools, called \dirty, is created.  We conduct an extensive experimental study on two widely used datasets using 14 classification models, revealing that performance degradation is not always predictable from simple correlations with the target variable. 
\keywords{Data Quality  \and Data Centric AI \and Machine Learning performance.}
\end{abstract}

\section{Introduction}

Multiple studies have explored the influence of data errors on the efficacy of machine learning (ML) models \cite{Qi2024,10.14778/3648160.3648178,Mohammed_2025}. These studies suggest that, due to the pronounced nonlinearity of datasets and models, it is challenging to generalize about how data quality affects model performance. Consequently, a dataset with certain errors might perform well with one machine learning model but poorly with another. Similarly, two datasets with the same error types, when used to train the same ML model, may yield significantly different results.  In this paper, we propose the Error Sensitivity Profile (ESP), a new method for assessing and comparing the impact of errors on ML performance. ESP is able to 1) detect for a given model trained on a dataset contaminated by a type of error applied to one or more features if there is a linear relationship between the percentage of errors and the performance degradation (Error Performance Correlation EPC), 2) calculate the performance that globally the model miss or gain with the increase of the error level (Area under Curve Error-Performance AEPC), and 3) the behavior of the model according to the different level of errors. With ESP, both practitioners and researchers can compare the effects of training data quality across models and datasets.
To fully exploit the ESP, we developed \textbf{\dirty} a suite of Python-based tools to aid specialists in examining the impact of specific dataset errors on the performance of specific \ml models. This is facilitated through a Python library named PuckTrick\cite{puctrick2025}, which is capable of injecting predefined percentages of specific errors. The analysis of two widely used datasets across 14 ML models demonstrates the effectiveness of ESP as a tool for assessing the impact of data quality on machine learning models. The study confirms the Dirtify suite's ability to support this type of research.  This paper is structured as follows: Section \ref{sec:soa} examines the current state of the art. Section \ref{sec:epc} introduces a formal definition of ESP, 
 while the tool suite and its key architectural components are outlined in Section \ref{sec:dirtify}. Section \ref{sec:casestudy} displays and analyzes the principal results from the empirical analysis. Lastly, Section \ref{sec:conclusion} concludes the discussion and outlines potential future works

\section{State of the Art}
\label{sec:soa}
Prior research has explored the impact of different types of dirty data on ML models, including missing values~\cite{Qi2024,10.1007/s00521-022-07702-7}, inconsistent data~\cite{Qi2024}, outliers~\cite{10912020},
label noise~\cite{6685834}, and categorical duplicates~\cite{10.14778/3648160.3648178}.
Frameworks for sensitivity analysis have been proposed
in~\cite{DBLP:conf/indin/DixMOBKSCMO23}, but are limited to a specific domain.
While these works addressed data quality issues in various ways, their analyses focused on a single error type; additionally, none have developed a flexible tool for dataset contamination and assessment comparable to \dirty. Recently, \cite{Mohammed_2025} investigated the impact of six data quality 
dimensions across multiple ML tasks, evaluating five classification algorithms 
(Logistic Regression, MLP, SVM, TabNet, and KNN) on a fixed collection of 
datasets. Although complementary in spirit, the two approaches differ along 
several dimensions. First, \cite{Mohammed_2025} operates at the dataset level, 
producing aggregate scalar indicators of quality impact, whereas ESP operates 
at the feature level and provides a multi-dimensional profile that jointly 
captures the direction, cumulative magnitude, and local structure of the 
error--performance relationship. Second, the quality dimensions adopted in 
\cite{Mohammed_2025} conflate structurally distinct error types, for instance, 
outliers and noisy values are subsumed under the single dimension feature accuracy, whereas ESP  treats each error type as a separate, independently controllable corruption  operator, enabling finer diagnostic resolution. Third, ESP is evaluated across  14 classification models, providing broader coverage of the algorithmic 
landscape. These differences make ESP and \cite{Mohammed_2025} complementary 
rather than competing: \cite{Mohammed_2025} offers a dataset-level view of 
quality impact across a curated benchmark, while ESP provides a 
fine-grained, model-specific sensitivity profile designed to guide 
feature-level data cleaning decisions.

CleanML~\cite{li2019cleanml} evaluates the impact of cleaning algorithms applied to real-world dirty datasets, using repeated runs and FDR control to ensure statistical rigour. The ESP profile can also be computed on top of experimental protocols such as CleanML, provided that the underlying methodology generates error--performance curves at controlled corruption levels.

Among tools for data corruption, BART~\cite{10.14778/2850578.2850579} allows users to specify error types and distributions via denial constraints, but operates only on relational tables and does not evaluate ML tasks. Jenga~\cite{schelter2021jenga} is a Python library that introduces data corruptions into datasets to evaluate model robustness, but tests a single model at a time and lacks a configuration interface for error strategies. The PuckTrick library~\cite{puctrick2025}, on which Dirtify relies, provides a broader range of error types than Jenga, including label perturbations and duplicated data. 
\section{Error Sensitivity Profile}
\label{sec:epc}
Let $D_0$ be a cleaned training dataset (our baseline) and let $\mathcal{T}_\theta : D \rightarrow D$
be the family of corruption operators, where $\theta$ is a vector of parameters
(e.g., error rate, error type, affected features).
For each error rate $k = 1, \ldots, K$, we define $D_k = \mathcal{T}_{\theta_k}(D_0)$ as a
corrupted training dataset.
Let $f_k = A(D_k; s_k)$ be a model based on algorithm $A$ trained on $D_k$ with random seed $s_k$, and let $p_k = \mathrm{Perf}(f_k, D_\mathrm{test})$ be a performance metric evaluated on a clean test set $D_\mathrm{test}$. We define a scalar measure of the training data degradation induced by $\mathcal{T}_{\theta_k}$ as $e_k = \mathrm{Err}_\tau(D_k, D_0)$, where $\tau$ denotes the error
type (e.g., label noise, missing values, outliers), and
$\mathrm{Err}_\tau : D \times D \rightarrow \mathbb{R}^+$ is an error-specific degradation function that quantifies the severity of the corruption with respect to the clean baseline $D_0$.
In the experimental evaluation presented in this paper,
$\mathrm{Err}_\tau(D_k, D_0)$ is instantiated as the percentage of corrupted values injected into the training set, i.e.\
$e_k \in \{0, 20, 40, 60, 80\}\%$, which provides a transparent and reproducible operationalisation of the corruption severity. 
Given a learning algorithm $A$, a baseline training dataset $D_0$, a corruption type $\tau$, and an associated family of corrupted datasets $\{D_k\}_{k=1}^{K}$, we define the \emph{Error Sensitivity Profile} of $A$ with respect to $\tau$ as the
tuple:
\begin{equation}
  \mathrm{ESP}(A, D_0, \tau) =
  \Bigl\langle\,
    \mathrm{EPC},\;
    \mathrm{AEPC},\;
    \{\hat{\beta}_j\}_{j=1}^{J}
  \,\Bigr\rangle,
\end{equation}
where each component isolates a specific, complementary facet of the
error--performance relationship, as detailed below.

\subsection{Error Performance Correlation}

The \emph{Error Performance Correlation} (EPC) provides a global, first-order summary of the relationship between data corruption and model performance. Specifically, EPC is defined as the negated Pearson correlation coefficient between the error severity sequence $\{e_k\}$ and the corresponding performance
sequence $\{p_k\}$:
\begin{equation}
  \mathrm{EPC} = -\,\rho\!\left(\{e_k\},\,\{p_k\}\right),
\end{equation}
where $\rho(\cdot,\cdot)$ denotes the Pearson correlation coefficient.
The negation is introduced so that a \emph{positive} EPC value indicates the expected inverse relationship between error level and model performance: as corruption increases, performance decreases.
Conversely, a negative EPC signals that performance improves as errors
accumulate, a counterintuitive but empirically observed phenomenon
(see Section~\ref{sec:casestudy}).
An EPC value close to zero indicates the absence of a globally linear
relationship between error and performance. 
It is important to note that EPC is intentionally designed as a  lightweight first-order diagnostic: it efficiently detects whether a linear trend dominates the error--performance curve, but it is  not intended to provide a complete characterisation on its own.
When EPC is close to zero, the relationship is non-linear or non-monotonic, and the remaining components of the ESP, AEPC and the piecewise slope vector, become the primary diagnostic tools.

\subsection{Area under the Error--Performance Curve}

The \emph{Area between the Error--Performance curves} (AEPC) quantifies the cumulative magnitude of the deviation from baseline performance across all error levels. To ensure comparability across models and datasets with different baseline performance levels, AEPC is defined as the normalised integral of the signed vertical distance between the baseline performance $p_0$ and the observed performance curve:
\begin{equation}
  \mathrm{AEPC} = \frac{\displaystyle\int_{0}^{e_{\max}}
    \bigl(p(e) - p_0\bigr)\,\mathrm{d}e}
    {p_0 \cdot e_{\max}},
    \label{eq:aepc}
\end{equation}
where $p(e)$ is the performance observed at error level $e$, $p_0$ is the baseline performance on the uncorrupted dataset, and $e_{\max}$ is the maximum corruption level considered.
The denominator $p_0 \cdot e_{\max}$ represents the area subtended by the baseline performance over the full corruption range, so that AEPC expresses the cumulative deviation as a fraction of the baseline contribution. A value of $\mathrm{AEPC} = -0.10$, for instance, indicates that the model lost, on average, $10\%$ of its baseline performance across all corruption levels.
It is worth noting that AEPC is meaningful only when computed with respect to a consistent performance metric: comparisons across tasks or metrics (e.g., F1 vs.\ accuracy) are not defined, and in the canonical representation, the metric adopted is always reported explicitly. Regarding numerical stability, the normalisation in  Equation~\ref{eq:aepc} is well-defined provided $p_0 > 0$. In the degenerate case $p_0 \approx 0$, the normalisation becomes numerically unstable; however, this scenario implies that the model achieves negligible baseline performance on the uncorrupted dataset, a condition under which the model itself is not a viable candidate for deployment, and a sensitivity analysis with respect to data quality would be of no practical relevance. We therefore consider this case outside the intended scope of ESP.
A positive AEPC indicates that, on average, the model performs better than the baseline across the corruption range, whereas a negative AEPC reflects a net degradation. Notice that, as with EPC, AEPC provides a global summary and may obscure local non-monotonic behaviour; the piecewise slope vector is therefore essential for a complete interpretation. While EPC captures the direction and consistency of the error--performance response, AEPC quantifies its overall relative extent, providing complementary information especially in non-monotonic regimes where EPC loses descriptive power.

\subsection{Piecewise Slope Vector}

To capture the local structure of the error--performance curve, we partition the error axis into $J$ locally monotonic regions and estimate a slope $\hat{\beta}_j$ within each region $j = 1, \ldots, J$.
The resulting \emph{piecewise slope vector} $\{\hat{\beta}_j\}_{j=1}^{J}$ reveals local regimes, critical thresholds, and saturation phenomena that neither EPC nor AEPC can express. Regime boundaries are identified by detecting sign changes in the first-order differences of the performance sequence $\{p_k\}_{k=1}^{K}$: a boundary is placed at position $k^*$ whenever
$\mathrm{sign}(p_{k^*} - p_{k^*-1}) \neq
\mathrm{sign}(p_{k^*+1} - p_{k^*})$.
Within each resulting monotonic region $\mathcal{R}_j =
\{k : k_j \leq k \leq k_{j+1}\}$, the slope $\hat{\beta}_j$ is estimated by ordinary least squares (OLS) regression of $p_k$ on
$e_k$:
\begin{equation}
  \hat{\beta}_j = \frac{
    \sum_{k \in \mathcal{R}_j}(e_k - \bar{e}_j)(p_k - \bar{p}_j)
  }{
    \sum_{k \in \mathcal{R}_j}(e_k - \bar{e}_j)^2
  },
\end{equation}
where $\bar{e}_j$ and $\bar{p}_j$ denote the mean error level and mean performance within region $\mathcal{R}_j$ respectively.
A negative $\hat{\beta}_j$ indicates performance degradation within that regime, while a positive value signals improvement.
\begin{figure}
    \centering
    \includegraphics[width=\linewidth]{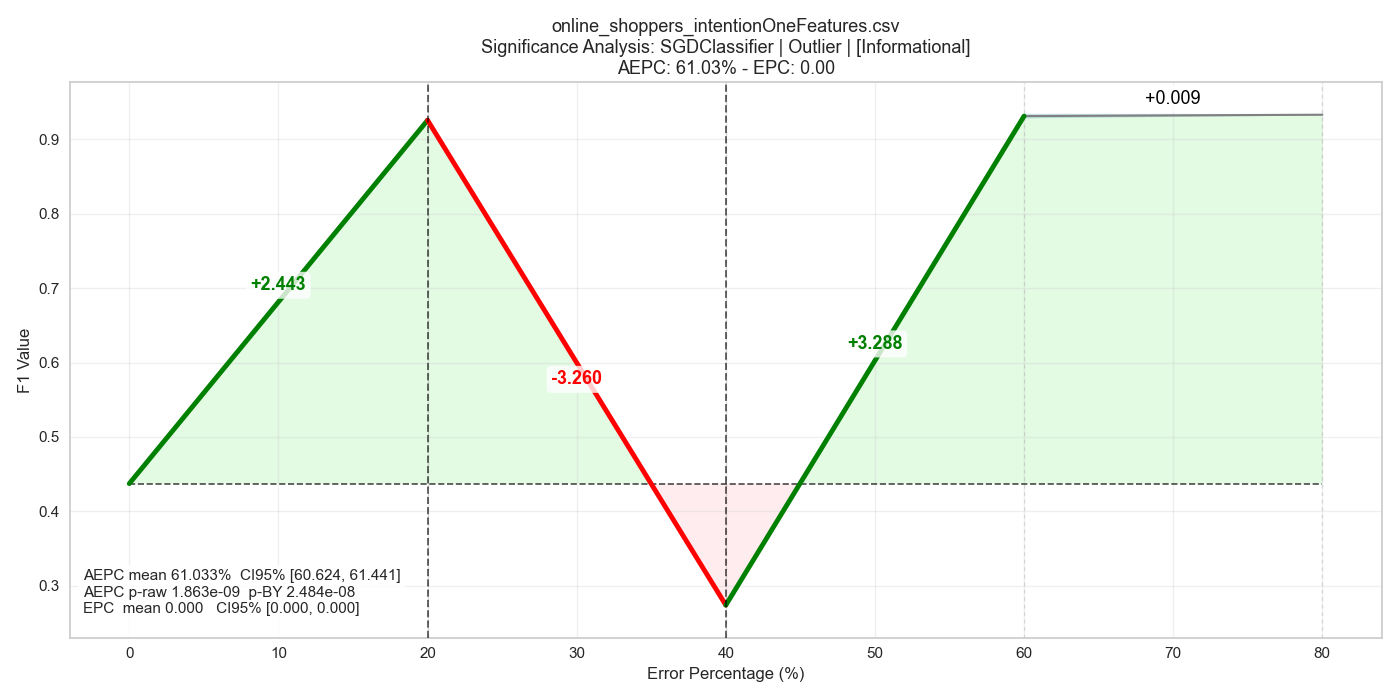}
    \caption{Canonical representation of ESP}
    \label{fig:ESPCanonical}
\end{figure}

It should be noted that, given the five corruption levels used in this study, slope estimates in short monotonic regions may be computed on as few as two observations, yielding exact fits with no residual degrees of freedom; such estimates should therefore be interpreted as directional indicators rather than statistically reliable regression coefficients, and increasing the number of corruption levels remains a direction for future work

\subsection{Canonical Representation}

To graphically represent and compare different ESPs, we adopt the canonical representation illustrated in Figure~\ref{fig:ESPCanonical}.
The performance--error curve is annotated with the global dependence (EPC), the cumulative degradation (AEPC), and the piecewise slopes identifying local regimes.

This unified visual form allows practitioners to assess, at a glance, both the overall sensitivity of a model to data quality issues and the structural features of its error--performance response. Since each experiment is repeated $N = 30$ times with different
random seeds (see Section~\ref{sec:casestudy}), in practice,  we compute an ESP for each independent run $n = 1, \ldots, N$ and report the aggregate ESP, whose components are defined as:
\begin{equation}
  \overline{\mathrm{EPC}} = \frac{1}{N}\sum_{n=1}^{N}\mathrm{EPC}^{(n)},
  \qquad
  \overline{\mathrm{AEPC}} = \frac{1}{N}\sum_{n=1}^{N}\mathrm{AEPC}^{(n)},
  \qquad
  \bar{\hat{\beta}}_j = \frac{1}{N}\sum_{n=1}^{N}\hat{\beta}_j^{(n)}.
\end{equation}
Confidence intervals at $95\%$ are reported alongside each aggregate component to quantify estimation uncertainty across runs, as shown in the canonical representation of Figure~\ref{fig:ESPCanonical}.
\section{Dirtify}
\label{sec:dirtify}

To support the computation of the ESP, we developed Dirtify, a suite of three loosely coupled Python-based tools. The \textit{Configurator}  is a wizard-style application that guides the user in defining an error  strategy and serialises it as a JSON configuration file. To support users in designing a strategy, the Configurator provides two predefined error strategies and allows users to define their own. The first  predefined strategy, referred to as \textit{one-feature-at-a-time}, applies each error type to each user-selected feature independently, enabling the user to isolate the contribution of each feature to model performance degradation. The second predefined strategy, referred to as \textit{correlated-features}, applies each error type to groups of features whose pairwise correlation exceeds a user-defined threshold, modelling the realistic scenario in which corrupting a single feature has limited impact due to redundancy with correlated features. Users may also define custom strategies by specifying arbitrary error  functions parametrised by error type, target features, corruption mode,  injection distribution, and a predicate for selecting the subset of  $D_0$ to be modified, enabling fine-grained control over which  instances are subject to corruption.

\begin{table}
    \centering
    \begin{tabular}{|l|l|l|l|}
         \hline
         Label & model Name &Label & model Name \\ 
         \hline
         SVM &  SVM - Linear Kernel    &NB &  Naive Bayes\\ 
         \hline
         ET & Extra Trees Classifier &          DT &  Decision Tree Classifier \\ 
         \hline
         RF & Random Forest Classifier &          QDA & Quadratic Discriminant Analysis   \\ 
         \hline
         KN &  K Neighbors Classifier  &        SGD & Stochastic Gradient Descent Classifier \\
         \hline
         LDA & Linear Discriminant Analysis &
         RC & Ridge Classifiers\\
         \hline
         MLP & Multilayer Perception Classifier &ADA & Ada Boost\\
         \hline
         LR  & Logistic Regression  &XG & XGBoost\\
         \hline

    \end{tabular}
    \caption{Machine learning algorithms used by the \dirty suite.}
    \label{tab:ml}
\end{table}

The \textit{Trainer} reads the JSON configuration produced by the Configurator and executes as follows: for each combination of error type, corrupted feature, and error level $e_k$, it corrupts the baseline dataset $D_0$ via the PuckTrick 
library~\cite{puctrick2025} and trains each selected ML model using PyCaret\footnote{\url{https://pycaret.org/}}, storing the resulting performance metrics. The precise operational definition of each error type, 
including injection distributions, parametrization, and categorical handling, is documented in the PuckTrick library~\cite{puctrick2025}. Each experimental configuration is repeated $N = 30$ times with different random seeds. For each repetition, the training/test split is generated once using an 80--20\% ratio with Stratified K-Folds cross-validation 
and held fixed across all corruption levels within that repetition. Crucially, corruption is applied exclusively to the training set: $D_{test}$ is never corrupted, ensuring that observed variations in $p_k$ are attributable solely to the degradation 
of training data quality and not to test-set resampling variance. The list of supported ML algorithms is shown in Table~\ref{tab:ml}. Hyperparameters are set to PyCaret defaults (see  Table~\ref{tab:param}) and held fixed across all corruption levels and repetitions, ensuring that observed performance variations are attributable solely to training data degradation and not to model recalibration.

\begin{table}
    \centering
    \begin{tabular}{|l|l|}
    \hline
Imputation type          &  simple \\
         \hline
Numeric imputation        &      means\\
         \hline
Categorical imputation     &         mode \\
         \hline
Fold Generator &  Stratified K-Folds\\
         \hline
Fold Number     &           10\\
         \hline
   \end{tabular}
    \caption{Hyperparameter and Configuration setting }
    \label{tab:param}
\end{table}

While fixing hyperparameters to their defaults ensures that observed performance variations are attributable solely to training data degradation, it also implies that models sensitive to configuration, such as MLP and XGBoost, may exhibit ESP profiles that partially reflect suboptimal parametrisation rather than intrinsic sensitivity to data quality, a confound that future work should address through hyperparameter-controlled experimental designs.

\section{Case study}
\label{sec:casestudy}
To illustrate how data quality impacts model performance and to demonstrate the insights provided by the proposed sensitivity profile and the Dirtify suite\footnote{\url{https://anonymous.4open.science/r/dirtify-4730}}, we consider the Online Shoppers Purchasing Intention dataset\cite{sakar2018online}, publicly available through the UCI Machine Learning Repository. The dataset comprises 12,330 sessions from distinct users, spanning 17 numerical and categorical features, including web analytics metrics such as bounce rates, exit rates, and page values, as well as contextual variables such as visit timing and visitor category, with no missing values and a well-defined binary target variable (\textit{Revenue}), making it well-suited for classification without extensive preprocessing.  Feature importance scores, computed using a Random Forest classifier (accuracy: 89.7\%), reveal a strongly skewed distribution: \textit{PageValues} dominates (0.383), followed by \textit{ProductRelated\_Duration} (0.088), \textit{ExitRates} (0.086), \textit{ProductRelated} (0.073), and \textit{BounceRates} (0.059), while demographic and technical attributes contribute negligibly. Pearson correlation analysis identifies two highly correlated feature pairs — \textit{BounceRates}/\textit{ExitRates} (r = 0.913) and \textit{ProductRelated}/\textit{ProductRelated\_Duration} (r = 0.861) — suggesting multicollinearity, while \textit{PageValues} shows the strongest association with \textit{Revenue} (|r| = 0.493). Given the significant class imbalance (84.5\% non-purchase vs. 15.5\% purchase), the F1-score was adopted as the primary evaluation metric. The one-feature-at-a-time strategy was applied, contaminating each feature independently with noisy values and outliers. The same error types were applied to correlated feature groups ($r \geq 0.5$),
following the correlated-features strategy. Mislabeling, duplication, and a custom oversampling strategy,  duplicating only rows where the
target equals 0,  were also investigated. In all strategies, corruption was applied incrementally from 0 to 80\% in steps of 20\%, with 30 independent runs per combination of model, error type, feature,
and corruption level. Overall, 21840 models were trained across 728 distinct scenarios, where a scenario is defined as a specific combination of error type, corrupted feature(s), and ML model.

\begin{figure}[h]
    \centering
    \includegraphics[width=.8\linewidth]{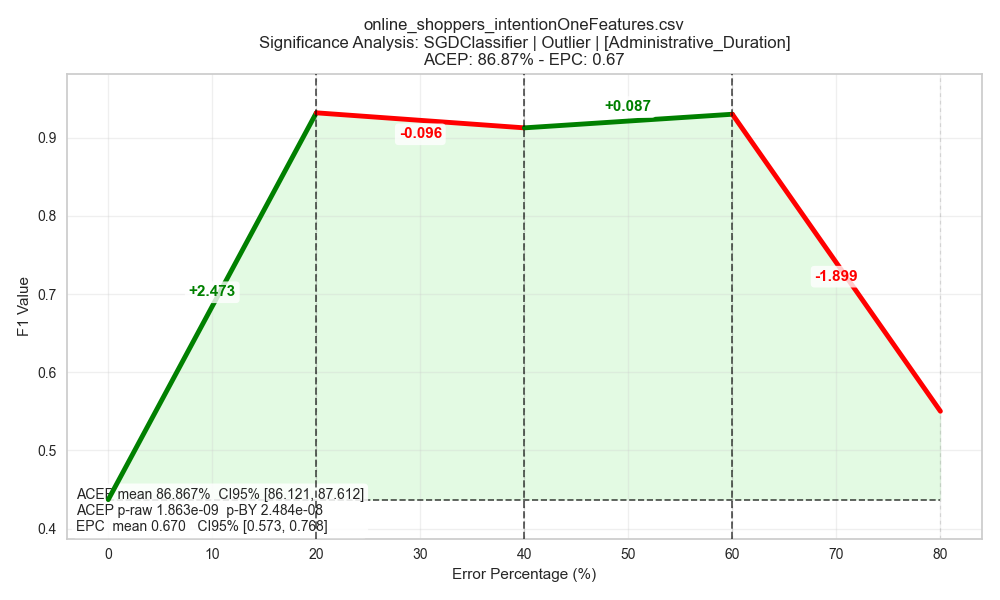}
    \caption{Relevant scenario where $\overline{\mathrm{AEPC}}$ is positive}
    \label{fig:scenario_3}
\end{figure}
\begin{figure}[h]
    \centering
    \includegraphics[width=\linewidth]{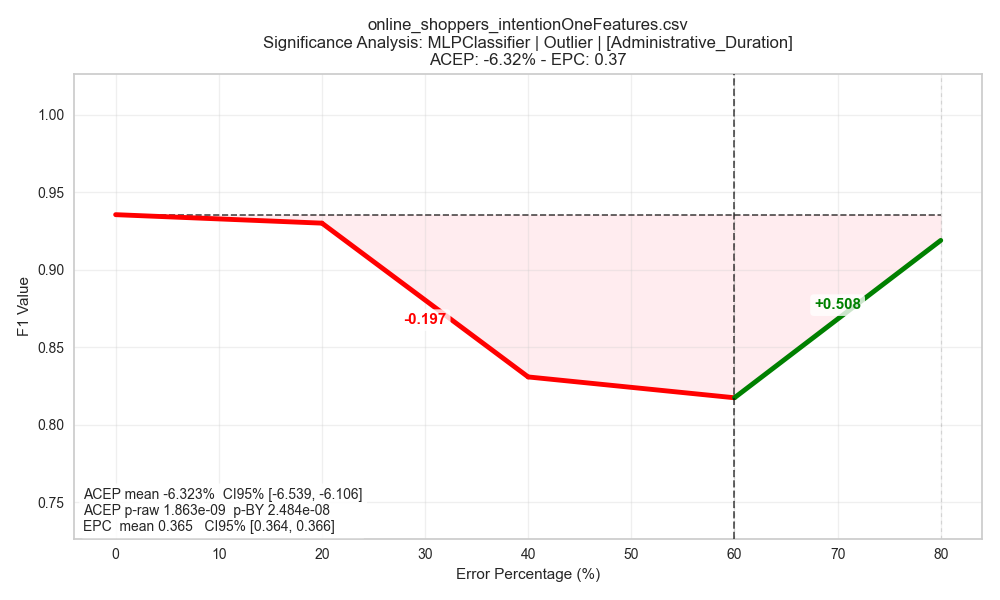}
    \caption{Relevant scenario where $\overline{\mathrm{AEPC}}$ is negative}
    \label{fig:scenario_2}
\end{figure}

Before analysing the ESP components, we applied a two-stage filtering procedure to identify scenarios that are both statistically and practically relevant, and to make the analysis tractable, given the large number of experimental configurations. In the first stage, statistical significance was assessed for each scenario independently. For each scenario, we compared the distribution of performance values obtained across the 30 independent runs at the maximum corruption level against the corresponding baseline distribution using a Wilcoxon signed-rank test, with a significance level of $\alpha = 0.05$. This non-parametric test was preferred over a standard $t$-test as it does not assume normality of the performance distributions, which cannot be guaranteed across all models and corruption types.To address the multiple testing problem arising from the simultaneous evaluation of all scenarios, raw p-values were corrected using the Benjamini-Yekutieli (BY) procedure~\cite{benjamini2001} at FDR level $\alpha = 0.05$, which provides FDR control under arbitrary dependence among tests. In the second stage, among the scenarios that passed the statistical significance criterion, we retained only those for which the absolute value of the normalised $\overline{\mathrm{AEPC}}$ exceeded a practical relevance threshold:
$
    |\overline{\mathrm{AEPC}}| > \delta, \quad \delta = 0.05.
$
This threshold corresponds to a cumulative deviation greater than $5\%$ of the baseline performance across the full corruption range.   Of the 728 scenarios, only 44, approximately 6\% of the total, were significant.
Deviations below this value were deemed negligible for practical data-cleaning purposes, as they would not meaningfully influence the decision about whether to invest resources in cleaning a given feature.
To verify the robustness of our findings with respect to this choice, we repeated the analysis with $\delta \in \{0.03, 0.10\}$, obtaining
$115$ and $36$ significant scenarios respectively
The qualitative findings reported below remain stable across all three thresholds, confirming that the choice of $\delta = 0.05$ does not introduce a material bias in the conclusions.
Among these scenarios, the model that appears most frequently, i.e., the one exhibiting the most significant changes, is the \textbf{SGD Classifier}, with 34 occurrences. Only two scenarios involve the correlated feature pair \textit{[ExitRates, BounceRates]}, and in both cases the model affected is the \textbf{Quadratic Discriminant Analysis}. Among the top four features by importance, the first two appear in 6 scenarios each; \textit{ExitRates}  appears only in one scenario, while \textit{ProductRelated} and \textit{BounceRates} each appear individually in 2 scenarios. It is worth noting that contamination of certain variables, such as \texttt{Month}, \texttt{Weekend}, and \texttt{Administrative\_Duration}, can significantly affect model performance, even though these variables are not among the most important features.  Considering $\overline{\mathrm{AEPC}}$, we observe that in 91\% of significant scenarios, the value is positive, meaning that in 40 out of 44 significant scenarios, corrupting the data increases model performance.
Among others, in the scenario shown in Figure \ref{fig:scenario_3}, we notices that the SGD trained on a dataset where the \texttt{Administrative\_Duration} is affected by outliers produces an  F1 score that improves sharply and stabilizes up to 60\% error rate, as confirmed by the strongly positive $\overline{AEPC}$ (86.87\%), indicating an overall beneficial effect of the outlier corruption on this model. However, beyond this threshold, the performance collapses dramatically. In the scenario of Figure \ref{fig:scenario_2}, where the MLP Classifier was trained on the same dataset used in the previous scenario, we observe that F1 exhibits behavior that is virtually specular to that of the previous one, even though the dataset is the same. In fact, a sharp performance drop up to 40\% error rate followed by a substantial recovery beyond the 60\% threshold, where the model progressively compensates for the introduced noise, almost fully restoring its baseline F1 score. 

To provide preliminary evidence of generalisability, we apply ESP to the South  German Credit dataset\footnote{\url{https://archive.ics.uci.edu/dataset/522/south+german+credit}}:  1,000 credit records with 20 features and a binary target (\texttt{kredit}),  moderate class imbalance (70/30\%), and dominant features \texttt{hoehe} (0.131)  and \texttt{laufkont} (0.116). Using the same experimental protocol, we obtained  980 scenarios, of which 77 ($\approx$8\%) were significant; \textbf{SGD} again dominates with 47 occurrences, and 86\% of significant scenarios exhibit a positive $\overline{\mathrm{AEPC}}$. Table~\ref{tab:comparison} summarises findings across both datasets: the vast majority of models exhibit remarkable robustness, while SGD accounts for 81 of 121 significant scenarios overall, a vulnerability attributable to its incremental update mechanism, which lacks the averaging effect of ensemble methods.

\begin{table}[h]
\centering
\caption{Comparative summary of ESP analysis across datasets.}
\label{tab:comparison}
\begin{tabular}{lcc}
\hline
 & Online Shoppers & South German Credit \\
\hline
Total scenarios          & 728      & 980 \\
Significant scenarios    & 44 (6\%) & 77 (8\%) \\
Most sensitive model     & SGD (34/44) & SGD (47/77) \\
Positive $\overline{\mathrm{AEPC}}$ (\%)       & 91\%     & 86\% \\
Top feature (importance) & PageValues (0.383) & hoehe (0.131) \\
Correlated pairs ($r \geq 0.5$) & 2 & 1 \\
\hline
\end{tabular}
\end{table}
Notably, low-importance features, such as
\texttt{Administrative\_Duration} and \texttt{TrafficType} in the first dataset can significantly destabilize SGD when heavily corrupted, suggesting that feature importance scores computed on clean data may underestimate the disruptive potential of marginal features under quality degradation. The error--performance relationship is rarely monotonic, as illustrated in Figures~\ref{fig:scenario_3} and~\ref{fig:scenario_2}, motivating the joint use of $\overline{\mathrm{AEPC}}$ and $\overline{\mathrm{EPC}}$ to capture both the magnitude and directionality of performance changes that neither metric alone could fully characterise. Furthermore, the counterintuitive prevalence of positive $\overline{\mathrm{AEPC}}$ values is confirmed on both datasets, in 91\% and 86\% of significant scenarios, respectively, suggesting that this phenomenon is not dataset-specific.
We conjecture that this phenomenon is partially attributable to class 
imbalance. Since the error types that produce significant scenarios 
involve corruptions applied exclusively to feature values rather than 
to class labels, the class distribution in the training set remains 
unchanged by construction. Nevertheless, injected outliers may alter 
the geometry of the feature space in ways that inadvertently improve 
the separability of minority-class instances, effectively acting as 
an implicit resampling mechanism. This hypothesis is consistent with 
the observed difference between the two datasets (91\% vs.\ 86\% of 
positive $\overline{\mathrm{AEPC}}$ scenarios), where the less imbalanced South German  Credit dataset exhibits a moderately lower prevalence of the effect,  though a formal verification is left to future work.

\section{Conclusion}
\label{sec:conclusion}
It is widely recognized that the quality of training data is crucial to the success of machine learning models. In this paper, we introduce the Error Sensitivity Profile and present Dirtify, an all-encompassing tool suite for systematically evaluating the impact of data quality on classification tasks. While \dirty facilitates a structured analysis of how various types of data degradation affect model performance, the Error Sensitivity Profile aids ML practitioners and researchers in determining the most robust model for a specific dataset and the resilience of ML models against data quality issues. We also present two significant applications of our approach by analysing well-known and complex datasets. 
The extensive experimental evaluation of about 51200 trained models (approximately 21800 for the first dataset and almost 29400 for the second) demonstrates the importance of empirically assessing the impact of data quality, rather than assuming that corruption invariably degrades performance.
While the experimental evaluation presented in this paper focuses on classification tasks, the ESP framework is designed to be task-agnostic: the performance function $\mathrm{Perf}(f_k, D_\mathrm{test})$ can be instantiated with any suitable metric, such as RMSE or MAE for regression, or silhouette score for clustering.
The extension of Dirtify to these settings is facilitated by the native support for regression and clustering pipelines in the underlying PyCaret library. Future work will extend the evaluation to a broader set of datasets; moreover we intend to expand Dirtify along four directions: introducing new error types, such as data drift, incorporating deep learning models, optimizing PuckTrick to scale to high-dimensional datasets, and investigating intelligent feature pre-screening strategies, based on importance scores, variance, or target correlation, to make the exhaustive ESP analysis tractable when hundreds of features are involved.
\bibliographystyle{splncs04}
\bibliography{sample}

\end{document}